\documentclass[a4paper,twoside]{article}

\usepackage{epsfig}
\usepackage{subcaption}
\usepackage{calc}
\usepackage{amssymb}
\usepackage{amstext}
\usepackage{amsmath}
\usepackage{amsthm}
\usepackage{multicol}
\usepackage{pslatex}
\usepackage{apalike}
\usepackage{SCITEPRESS}     

\usepackage{hyperref} 
\usepackage{algorithmic} 
\usepackage{algorithm} 

\begin{document}

\title{Adversarial Unsupervised Domain Adaptation Guided with Deep Clustering for Face Presentation Attack Detection}


\author{\authorname{Yomna Safaa El-Din, Mohamed N. Moustafa and Hani Mahdi}
	\affiliation{\sup{1}Computer and Systems Engineering Department, Ain Shams University, Cairo, Egypt}
	\affiliation{\sup{2}Department of Computer Science and Engineering, The American University in Cairo, New Cairo, Egypt}
	\email{\{yomna.safaa-eldin,~hani.mahdi\}@eng.asu.edu.eg, m.moustafa@aucegypt.edu}
}


\abstract{
    Face Presentation Attack Detection (PAD) has drawn increasing attentions to secure the face recognition systems that are widely used in many applications. Conventional face anti-spoofing methods have been proposed, assuming that testing is from the same domain used for training, and so cannot generalize well on unseen attack scenarios. The trained models tend to overfit to the acquisition sensors and attack types available in the training data. 
    In light of this, we propose an end-to-end learning framework based on Domain Adaptation (DA) to improve PAD generalization capability. 
    Labeled source-domain samples are used to train the feature extractor and classifier via cross-entropy loss, while unsupervised data from the target domain are utilized in adversarial DA approach causing the model to learn domain-invariant features. 
    Using DA alone in face PAD fails to adapt well to target domain that is acquired in different conditions with different devices and attack types than the source domain. And so, in order to keep the intrinsic properties of the target domain, deep clustering of target samples is performed. 
    Training and deep clustering are performed end-to-end, and experiments performed on several public benchmark datasets validate 
    that our 
    proposed Deep Clustering guided Unsupervised Domain Adaptation (DCDA) can learn more 
    generalized information compared with the state-of-the-art classification error on the target domain.
}


\onecolumn \maketitle \normalsize \setcounter{footnote}{0} \vfill

\section{\uppercase{Introduction}}

Face detection and recognition is an important topic in computer vision, it is used in many applications from which authentication is the most sensitive. Since the wide spread of smart mobile devices and the incorporation of latest vision technologies in these devices, end users find it more convenient to use their biometric data for authentication instead of classic passwords typing. On the other hand, this ease of use makes it easier for attacker to spoof the authentication system using pre-recorded biometric samples of the device user. Hence, the interest in developing reliable anti-spoofing or Presentation Attack Detection (PAD) techniques is increasing. Through the past years, several approaches were developed in literature~\cite{iet_yomna} starting from basic methods relying on image processing and hand-engineered features, till approaches depending on automatically learnt features by deep-learning.

These approaches have succeeded to obtain perfect attack detection results on intra-dataset scenarios, where the dataset is split into training and testing subsets, so both subsets are coming from the same sensor model and acquisition environment. 
However, the main drawback of such methods is their lack of generalization to different environments and attack scenarios. The performance of the learnt representations in classifying the attack from the bona-fide (real) presentation degrades significantly when test data is captured by different sensor or in different settings or illumination conditions. In view of this, Domain Adaptation (DA)~\cite{dann_2_2016} and Domain Generalization (DG)~\cite{MMD_AAE} were introduced recently in the PAD field. The target of DG is to learn representations that are robust across different domains, given samples from several source domains, such as in~\cite{learn_gen_DFR_face_2018}, \cite{MADDG_2019}, \cite{Jia_2020_CVPR_SSDG}. While, DA aims at adapting a model trained on labeled source domain to a different target domain. Unsupervised DA (UDA) uses labeled samples from a source domain and unlabeled samples from a target domain, with a goal to achieve low classification error on the target domain though samples are unlabeled, by learning domain-invariant features.

For example,~\cite{UDA_face_journal_2018} experimented with both hand-crafted and deep learnt features in DA, however their approach was not end-to-end and the deep features did not generalize well. They achieved their best results using a combination of hand-crafted features. Adversarial training was used in DA for face PAD in~\cite{impr_face_ADA_2019} to learn an embedding space shared by both the source and target domain models. The training process is still not end-to-end where source pre-training, embedding adaptation and target classification are done separately. 

In this paper, we focus on developing an end-to-end trainable solution for PAD based on DA, which focuses on improving the generalization of the model for cross-dataset testing without the need for several labeled source domains as in DG. Existing DA-based solutions solely aim to align the distribution of an unlabeled target domain to that of a different source domain, neglecting the specific nature of target domain. Target domain in face PAD is a different PAD dataset usually using a different device for authentication, in addition to different attack types in different illumination conditions. So solely trying to align the distribution of such different attacks scenarios to the distribution of attack scenarios in the labeled source dataset would not succeed, especially when the device used for authentication in one domain, is close to the one used for attack in the other domain, e.g.\ mobile device. So, we propose an approach that utilizes DA for PAD generalization to a different domain without neglecting the intrinsic properties of this target domain. We incorporate clustering based on deeply extracted features, for guiding the feature extraction network to generate features that are domain invariant, yet maintain the class-wise separability of the target dataset.

The main contributions of this work are: 
(1) proposing a novel end-to-end DA-based training architecture for the generalization of face PAD based; (2) utilize deep embedding clustering of target domain in guiding the DA process; (3) show substantial improvement on SOTA in cross-dataset evaluation on public benchmark face PAD datasets, with close to 0\% cross-dataset error. The rest of the paper is organized as follows: Section~\ref{sec:related} reviews the latest literature in face PAD and domain adaptation. Our proposed algorithm is explained in Section~\ref{sec:proposed}, 
followed by the experiments, benchmark datasets used and results in 
Section~\ref{sec:experiments}, then conclusions in Section~\ref{sec:conc}.

\section{\uppercase{Related work}}\label{sec:related}

\subsection{CNN-Based Face PAD}
Recent software-based face presentation attack detection methods can be mainly categorized into texture-based and temporal-based techniques. The texture-based methods rely on extracting features from the frames that would identify if the presented image is fake or bona-fide. Features could be hand-crafted features as 
color texture~\cite{545_16_f}, SIFT~\cite{553_16_f} or SURF~\cite{546_17_f} 
which obtained good results in differentiating real from fake presentations. However, 
they 
are often sensitive to varying acquisition conditions, such as camera devices, lighting conditions and Presentation Attack Instruments (PAIs). 
Hence, the need to automatically learn and extract meaningful features directly from the data using deep representations, such as in~\cite{317_18_f,iet_yomna}. 

In additional to texture-based features, temporal-based models utilize the temporal information in face videos for better detection of attack presentations. Frame difference was combined with deep features in~\cite{318_16_f}. In~\cite{311_16_f} image quality information and motion information from optical flow were combined with neural network for classification. LSTM-CNN architecture was used in~\cite{310_15_f} and in~\cite{302_18_f} multiple RGB frames were used to estimate face depth information, and then two modules were used to extract short and long-term motion.

These methods obtain excellent results in intra-dataset testing, yet still fail to generalize to unseen environments and acquisition conditions. They show high cross-dataset evaluation errors, hence the need to incorporate domain adaptation techniques to decrease the discrepancy in distributions of the domain used for training and that used for deployment.

\subsection{Unsupervised Domain Adaptation}

Recently, Domain Adaptation (DA) has been introduced in 
computer vision, to tackle the problem of domain shift when applying models trained on a certain (source) domain to another (target) domain. 
Several methods, such as~\cite{dann_2_2016}, rely on adversarial training~\cite{GAN_2014} to guide the feature extraction module to generate domain-invariant features that make it harder for a domain discriminator to decide the original domain of the sample. 
Specifically, unsupervised DA uses labeled samples from the source domain in addition to unlabeled samples from the target domain; to train a model that reduces the classification error on the unlabeled target domain. 

Inspired by the success of DA in image classification~\cite{MADA_2018}, \cite{Long_CDAN_2018}, \cite{MCD_Saito_2018}, \cite{Saito_2018}, \cite{Kurmi_2019}, \cite{Symnet_2019}, \cite{DADA_2020}, 
\cite{CAN_2020}, we believe that 
it can be used to address the problem of generalization in face PAD. 
A model fine-tuned on certain small-sized face PAD dataset fails to generalize when testing on different PAD domains with different domain. 
The learnt features become specific to the subjects or sensors 
available in the source dataset. Hence, 
by using domain adaptation in face PAD
, the model will be guided 
to learn domain-invariant features that can differentiate between bona-fide and attack face videos regardless of the instance origin. 
However, learning domain invariant features can hurt classification of the target face PAD dataset by ignoring the fine-level class-wise structure of this target since the attack samples are generated with different instruments, and bona-fide samples may be captured by different sensors. Hence, we propose to incorporate deep clustering of target samples to constraint the model to keep the discriminative structure of both classes in the target dataset.

\subsection{Deep Unsupervised Clustering}

Deep learning is adopted in clustering of deep visual features since Deep Embedded Clustering (DEC)~\cite{DEC}. Clustering aims at categorizing unlabeled data into groups (clusters). A DEC is a method that jointly learns feature representations and cluster assignments, 
where a neural network is first pre-trained by means of an autoencoder and then fine-tuned by jointly optimizing cluster centroids in output space and the underlying feature representation using Kullback-Leibler divergence minimization.
Later, variants of DEC have emerged, such as~\cite{DEC_with_aug18} which adds data augmentation.

Unlike DEC, which require layer-wise pretraining as well as non-joint embedding and clustering learning, \textbf{DE}e\textbf{P} Embedded Regular\textbf{I}zed \textbf{C}lus\textbf{T}ering 
(DEPICT)~\cite{DEPICT} utilizes an end-to-end optimization for training all network layers simultaneously using the unified clustering and reconstruction loss functions. 
DEPICT consists of a multi-layer convolutional autoencoder followed by a multinomial logistic regression function. 
The clustering objective function 
uses relative entropy (KL divergence) minimization, regularized by a prior for the frequency of cluster assignments. An alternating strategy is then 
followed to optimize the objective by updating
parameters and estimating cluster assignments. 
Reconstruction loss functions is employed in the
autoencoder 
to prevent the deep embedding function from overfitting. 
A joint learning framework is introduced to minimize the unified clustering and reconstruction loss functions together and train all network layers simultaneously.


Recently, clustering has been introduced in several domain adaptation methods. 
\cite{Wang2019DiscriminativeCF} 
proposed a method to alleviate the effects of negative transfer in adversarial domain matching between source and target representations. They proposed to 
simultaneously learn 
tightly clustered target representations while encouraging that each cluster is assigned to a unique and different class from the source. 
In~\cite{SRDC_2020}, structural domain similarity is assumed and the clustering solution is constrained using structural source regularization. By 
minimizing the KL divergence between predictive label distribution of the network and an introduced auxiliary one; replacing the auxiliary distribution with that formed by ground-truth labels of source data implements the structural source regularization via a simple strategy of joint network training. 

\begin{figure*}[tph]
    \centering
    \includegraphics[width=0.98\linewidth]
    {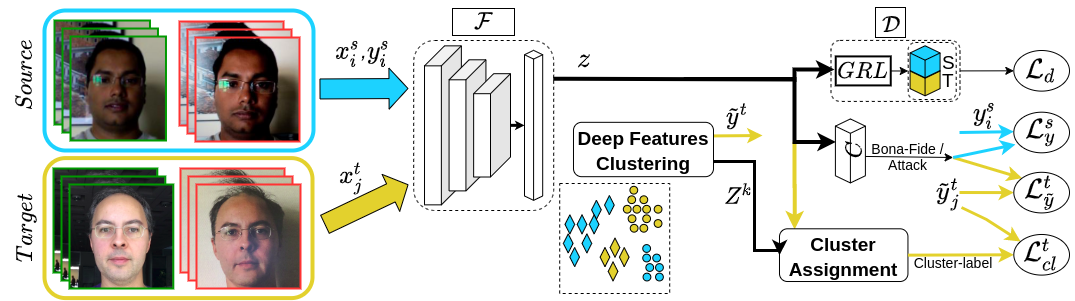}
    \caption{Architecture of the proposed Deep Clustering-guided-Domain Adaptation (DCDA) for face PAD. $\mathcal{F}$: Feature extraction network, $\mathcal{D}$: Domain Discriminator, $GRL$: Gradient Reverse Layer, $\mathcal{C}$: Categories Classifier, $S$: Source, $T$: Target. Bona-fide images are highlighted in green border, while attack images are highlighted in red. \textbf{Deep Features Clustering}: predicts target pseudo-labels $\tilde{y}$ and cluster centers $Z^k$. \textbf{Cluster Assignment}: assigns target features to clusters based on Student's $t$-distribution.}
    \label{fig:oct2020-clust-paper-28oct2020}
\end{figure*}

\subsection{DA in Face PAD}

Domain Adaptation (DA) and Domain Generalization (DG) have been utilized recently to reduce the gap between the target domain and the source domain during face PAD. 
\cite{MADDG_2019} focuses on improving the generalization ability of face PAD methods from the perspective of the domain generalization. Adversarial learning was proposed to train multiple feature extractors to learn a generalized feature space. 
They also incorporated an auxiliary face depth supervision to further enhance the generalization ability. 
Later, a Single-Side Domain Generalization framework was proposed in (SSDG)~\cite{Jia_2020_CVPR_SSDG} that is end-to-end. They proposed to learn a generalized feature space, where the feature distribution of the real faces is compact while that of the fake ones is dispersed among domains but compact within each domain.

One of the first work exploring DA for face PAD is~\cite{UDA_face_journal_2018} were both hand-crafted features and deep neural network learned features are adopted and compared in DA. 
\cite{UDA_face_journal_2018} found that the deep learning based methods may not generalize well under cross-database testing scenarios, and their best results were achieved using concatenated CoALBP and LPQ feature in HSV and YCbCr color space.

A 3D CNN architecture tailored for the spatial-temporal input is proposed by 
\cite{learn_gen_DFR_face_2018} for enhancing the generalization capability of the network. A robust representation across different face spoofing domains is presented by introducing the generalization loss as the regularization term. Given training samples from several domains, the network is optimized such that the Maximum Mean Discrepancy (MMD) distances among different domains can be minimized. They performed the experiments by combining three publicly available face PAD datasets 
to create 10 protocols. In each protocol, data from one camera is set aside as the unseen target domain, and a subset of the remaining cameras are used as source domains.

ADA~\cite{impr_face_ADA_2019} is the first to incorporate adversarial domain adaptation in a 
learning approach to improve face PAD generalization capability. 
A source model optimized with triplet loss is first pre-trained in source domain, and then adversarial adaptation is used for training a target model to learn a shared embedding space by both the source and target domain models. 
Finally, target images are mapped with the target model to the embedding space and classified with k-nearest neighbors' classifier. However, as the first attempt to use adversarial training for domain adaptation, the training is not performed end-to-end.
In~\cite{DA_face_mobile_2020}, 
authors relied only on bona-fide samples of the target domain for DA. They hypothesize that, in a CNN trained for PAD given a source domain,
some of the filters learned in the initial layers are robust filters that generalize well to the target dataset, whereas others are more specific to the source dataset. They propose to prune such filters that do not generalize well from one dataset to another in order to improve the performance of the network on the target dataset. Feature Divergence Measure (FDM) is computed to quantify the level of domain shift at a given layer in a CNN.

\cite{cross_dom_face_DA_2020} proposed disentangled representation learning for cross-domain face PAD. Their approach consists of Disentangled Representation learning (DR-Net) and Multi-Domain feature learning (MD-Net). DR-Net learns a pair of encoders via generative models that can disentangle PAD informative features from subject discriminative features. The disentangled features from different domains are fed to MD-Net which learns domain-independent features for the final cross-domain face PAD task. They tested single-source to single-target cross-domain PAD and also  multi-source to multi-target and obtained state of the art results on four public datasets. 
Their later work (DR-UDA)~\cite{UDA_cross_dom_face_journal_2021} 
consists of three modules, ML-Net, UDA-Net and DR-Net. ML-Net uses the labeled source domain face images to learn a discriminative feature representation. UDA-Net performs unsupervised adversarial domain adaptation in order to optimize the source domain and target domain encoders jointly, and obtain a common feature space shared by both domains. 
Furthermore, DR-Net disentangles the features irrelevant to specific domains by reconstructing the source and target domain face images from the common feature space.

\section{\uppercase{Methodology}} 
\label{sec:proposed}

In this section, we introduce the frameworks of unsupervised DA and unsupervised clustering. Then, we present our proposed model for UDA in face PAD. Figure~\ref{fig:oct2020-clust-paper-28oct2020} shows a brief overview of the proposed architecture.

Since the most common target platform is mobile devices, we follow~\cite{icmv_yomna_20} and use latest architecture of MobileNet; MobileNetV3~\cite{MobileNetV3_2019} instead of the commonly used Resnet-50~\cite{Resnet}. 
MobileNet is tuned for mobile phone CPUs which helps preserve the mobile battery life by reducing power consumption. With $\sim 80\%$ less parameters, MobileNetV3 achieves comparable ImageNet accuracy as Resnet50 with reduced inference time.

\subsection{Deep Unsupervised Domain Adaptation}

Unsupervised Domain Adaptation (UDA), depends on having a set of labeled source samples $S=\{(x_i,y_i)\}^{N_s}_{i=1}$ and another set of unlabeled samples from target domain $T=\{(x_j)\}^{N_t}_{j=1}$. The goal is to train a model that is capable of achieving low classification errors on the unlabeled target domain guided by the labeled source samples. The feature extraction module is trained to be able to extract features that benefit the categories classification without differentiating the domain origin of the sample.

As 
(DANN)
~\cite{dann_2_2016},
adversarial training is incorporated to guide the feature extraction module, $\mathcal{F}$, to generate features that confuse a domain discriminator, $\mathcal{D}$, to not be able to determine the domain of the input features.
The categories (task) classifier, $\mathcal{C}$, is then trained on top of these generated domain-invariant features; using the labeled source samples, to decide the final classification label.

The task classification loss is calculated as
\begin{equation}  \label{Loss:class_s}
    L^s_y = \frac{1}{N_s}\sum_{i=1}^{N_s} \mathcal{L}_{y} (\mathcal{C}(\mathcal{F}(x_i)), y_i),
\end{equation}
where $\mathcal{L}_{y}$ is categorical cross-entropy loss, 
$\mathcal{F}$ is the feature extractor network and $L^s_{y}$ is the task classification loss from all source samples using. Similarly, domain discrimination loss, 
\begin{equation}  \label{Loss:dom}
    L_d = \frac{1}{N_s+N_t}\sum_{m=1}^{(N_s+N_t)} \mathcal{L}_{d} (\mathcal{D}(\mathcal{F}(x_m)), d_m),
\end{equation}
where $\mathcal{L}_{d}$ is categorical cross-entropy loss, $d_m$ is domain label, zero for source samples, and one otherwise. This loss is minimized over the parameters of $\mathcal{•}{D}$ while maximized over the parameters of $\mathcal{F}$ via the gradient reverse layer ($GRL$).

\subsection{Proposed DC-guided UDA for Face PAD}

For handling the problem of generalization in face PAD, we propose to use UDA, in combination with Deep Embedding Clustering (DEC) of the unlabeled target samples during training. Motivation for UDA is to alleviate the shift between the source and target domains.
However, we do not want to lose the target properties for each class. 

Aligning both source and target domains in face PAD with source and target coming from different sensors and attack instruments, might lead to target samples being misclassified and shifted towards the wrong class. For example, a target mobile attack instance can be assigned to the closest source sample which might be bona-fide class if bona-fide samples of source dataset are captured with same instrument (mobile device). 
So motivation for adding target clustering is to preserve 
the class-wise separation of target domain samples. Which together with adversarial DA, will guide $\mathcal{F}$ to generate features that reduce domain shift without corrupting the class-wise separability of target domain.

\begin{algorithm}[] 
    \caption{Training of DCDA: Deep Clustering-guided-Domain adaptation for face PAD}
    \label{alg:alg}
    \begin{algorithmic}
        \STATE Let \{$\theta_{\mathcal{F}}$, $\theta_{\mathcal{C}}$, $\theta_{\mathcal{D}}$\} be the learnable parameters for each model component.
        \STATE Let \{$Z^k_{BF}$ , $Z^k_A$\} be the learnable cluster centers for bona-fide and attack classes respectively.
        
        \STATE {\textbf{Input:}} 
        \STATE \hspace{3mm} Labeled source videos $S: (X^s,Y^s)$ and unlabeled target videos $T: (X^t)$ 
        \STATE \hspace{3mm} Batch size: $B$
        \STATE {\bfseries Output:}
        \STATE \hspace{3mm} Feature extractor: $\mathcal{F}(\cdot)$
        \STATE \hspace{3mm} Classifier: $C(\cdot)$
        
        \STATE {\textbf{Deep Descriminative Clustering:}} 
        \STATE \hspace{3mm} Fix model parameters
        \STATE \hspace{3mm} $\{z^s_i\} = \mathcal{F}(x^s_i)$ for all $x^s_i \in X^s$
        \STATE \hspace{3mm} $\{z^t_j\} = \mathcal{F}(x^t_j)$ for all $x^t_j \in X^t$
        \STATE \hspace{3mm} $Z^k_{c^s} = avg(\{z^s_i\})$ for $y^s_i = c$ $\forall$ $c \in \{BF, A\}$
        \STATE \hspace{3mm} $\tilde{y}^t, Z^k_{BF^t}, Z^k_{A^t}$ $\leftarrow$ k-means clustering of $\{z^t_j\}$ using $Z^k_{BF^s}, Z^k_{A^s}$ as initial centers 
        \STATE \hspace{3mm} \textbf{for} $c \in \{BF, A\}$ \textbf{do}
        \STATE \hspace{6mm} $Z^k_{c^s} = avg(\{z^s_i\})$ for $y^s_i = c$
        \STATE \hspace{6mm} $Z^k_c = avg(Z^k_{c^s},Z^k_{c^t})$
        
        \STATE $ep=0$
        \WHILE {$ep < max\_epochs$} 
        
        \FOR{$b=0$ \TO $iter\_per\_epoch$}
        \STATE Draw random batch $\{(x^s_i,y^s_i)\}^{B}_{i=1}$, $\{(x^t_j,\tilde{y}^t_j)\}^{B}_{j=1}$
        \STATE $\theta_{\mathcal{F}} = \theta_{\mathcal{F}} -  \nabla_{\theta_{\mathcal{F}}}( L_y^s + L_{\tilde{y}}^t + L_{d} + L_{cl}^t)$
        \STATE $\theta_{\mathcal{C}} = \theta_{\mathcal{C}} -  \nabla_{\theta_{\mathcal{C}}} (L_y^s
        + L_{\tilde{y}}^t)$
        \STATE $\theta_{\mathcal{D}} = \theta_{\mathcal{D}} -  \nabla_{\theta_{\mathcal{D}}} L_d$
        \STATE $Z^k_c = Z^k_c - \nabla_{Z^k_c} L_{cl}^t , \forall c \in \{BF, A\}$
        \ENDFOR
        \STATE Update target pseudo-labels $\tilde{y}^t$ based on $\{z^t_j\}$ distance to $Z^k_{BF}$ and $Z^k_{A}$
        \STATE $ep = ep + 1$
        \ENDWHILE
    \end{algorithmic}
\end{algorithm}

\subsubsection{Deep Clustering for DA}

Our training follows the unsupervised deep clustering methods~\cite{DEC}, \cite{DEPICT} which alternates between cluster assignment while fixing model parameters, then model update while fixing these cluster assignment. 
At the start of each epoch, k-means clustering is performed on the deep features generated by $\mathcal{F}$ to generate pseudo-labels, $\tilde{Y}^t$, for the unlabeled target samples. Then, during epoch iterations, two losses based on Kullback-Leibler (KL) divergence~\cite{DEC} are minimized 
to update the parameters of $\mathcal{F}$, $\mathcal{C}$ and cluster centroids $Z^k$ via back-propagation.

These learnable centroids $Z^k=\{Z^k_{BF}, Z^k_A\}$ for each of the bona-fide and attack classes are re-updated at the start of each epoch, while fixing the model parameters. Guided by the labels of source samples, and the source features generated by the current $\mathcal{F}$, clusters centers for the source domain; $Z^k_{c^s}$, can be obtained in the embedding space. On the other hand, for the unlabeled target samples, k-means clustering is used on the generated latent features of all target samples. This obtains both pseudo-labels for all target instances in training, $\tilde{Y}^t$, and clusters centers for the target domain, $Z^k_{c^t}$. 
Finally, 
the learnable cluster center for each class $Z^k_c$ is updated to be the mean of both $Z^k_{c^t}$ and $Z^k_{c^s}$.

During training iterations of an epoch, 
target samples are used to minimize KL divergence two-way. 
The loss to be minimized can be written as

\begin{align} \label{Loss:dec}
    L_{dec} &= KL(Q||P) + L_{reg}\\
    & = \frac{1}{N}\sum_{j=1}^{N}\sum_{k=1}^{K} q_{jk}\log\frac{q_{jk}}{p_{jk}}  + \sum_{k=1}^{K} \hat{q}_{k} \log \hat{q}_{k} \,,
    \nonumber
\end{align}
where $P^t$ is the cluster assignments for target samples and $Q^t$ is an auxiliary target distributions, and the purpose of Kl divergence minimization is to decrease the distance between the model predicted $P^t$ and the distribution $Q^t$. 
The second term follows~\cite{DC2010} for incorporating class balance to avoid degenerate solutions, where $\hat{q}_{k}=\frac{1}{N_t}\sum_{j=1}^{N_t} q^t_{jk}$.

As in~\cite{DEPICT}, optimization of loss in equation~\ref{Loss:dec} alternates between updating auxiliary distribution $Q^t$ then using $Q^t$ to update model parameters. $Q^t$ is calculated in closed-form solutions as

\begin{align} \label{eq:q}
    &	 q^t_{jk} =   \frac{  p^t_{jk}   / (\sum_{j'} p^t_{j'k})^{\frac{1}{2}}  }{\sum\limits_{k'}   p^t_{jk'}   / (\sum_{j'} p^t_{j'k'})^{\frac{1}{2}}}   \,.
\end{align} 
For further regulation of target clustering, we use the previously estimated target pseudo-labels as part of $Q^t$ by setting $q^t_j = 0.5 * q^t_j + 0.5 * \tilde{y}^t_j$.

Then using calculated $P^t$ and $Q^t$, parameters of $\mathcal{F}$ and $\mathcal{C}$ are updated by minimizing
\begin{equation} \label{eq:loss_kl_t_1}
    L^t_{cl} = -\frac{1}{N_t} \sum\limits_{j=1}^{N_t} \sum\limits_{k=1}^K q^t_{jk} \log  p^t_{jk} \,,
\end{equation}

As mentioned earlier, we use KL divergence minimization with target domain samples for two losses which update parameters of feature extraction module $\mathcal{F}$ via backpropagation. 
The first loss additionally aims to update the classifier $\mathcal{C}$ as well, and the second loss updates the cluster centroids $Z^k$. For the first loss ($L^t_{\tilde{y}}$), we 
set $P^t$ as the classifier prediction probabilities after softmax; $p^t_j = softmax(\mathcal{C}(\mathcal{F}(x^t_j)))$, so that it becomes like cross-entropy classification loss using pseudo-labeled target samples. 

For the second loss ($L^t_{cl}$), 
$P^t$ is estimated using the Student's \emph{t}-distribution 
to measure the similarity between target features $Z^t$ and cluster centroids $Z^k$ as in~\cite{DEC}
\begin{equation*}
    p^t_{jc}=\frac{(1+||z^t_j-Z^k_c||^2/\alpha)^{ -\frac{\alpha+1}{2} }}{ \sum_{ c^\prime }(1+||z^t_j-Z^k_{c^\prime}||^2/\alpha)^{-\frac{\alpha+1}{2}} } \,.
\end{equation*}

Finally, the estimated pseudo-labels for target samples are used to update the parameters of both the feature extractor $\mathcal{F}$ and the classifier $\mathcal{C}$ by minimizing the following task classification loss
\begin{equation}  \label{Loss:class_t}
	L_{\tilde{y}}^t = \frac{1}{N_t}\sum_{j=1}^{N_t} \mathcal{L}_{\tilde{y}} (\mathcal{C}(\mathcal{F}(x_j)), \tilde{y}_j),
\end{equation}
where $\mathcal{L}_{\tilde{y}}$ is categorical cross-entropy loss.

\subsubsection{Complete Model learning}

The complete end-to-end training methodology of our proposed DC-guided-DA for face PAD is listed in Algorithm~\ref{alg:alg}. We use only 
one frame per video.

\begin{table*}[t] %
    \caption{Number of samples per class per subset for each used PAD dataset.}
    \label{tab:face-db-numbers}
    \begin{center}
        \small
        
        \begin{tabular}{|l|l|l|l|ll|l|}
            \hline
            Database      & PAI 
            & Sensor used for authentication   & Subset                                                       & Bona-fide                                               & Attack                                                  & Total                                                   \\
            \hline
            Replay-Attack & \begin{tabular}[c]{@{}l@{}}1) \textbf{PR} (A4)\\ 2) \textbf{VR} on iPhone\\ 3) \textbf{VR} on iPad\end{tabular}              & (1) 
            Webcam in MacBook laptop                                                                                                                                         & \begin{tabular}[c]{@{}l@{}}train\\ devel\\ test\end{tabular} & \begin{tabular}[c]{@{}l@{}}300\\ 300\\ 400\end{tabular} & \begin{tabular}[c]{@{}l@{}}60\\ 60\\ 80\end{tabular}    & \begin{tabular}[c]{@{}l@{}}360\\ 360\\ 480\end{tabular} \\
            
            \hline
            MSU-MFSD      & \begin{tabular}[c]{@{}l@{}}1) \textbf{PR} (A3)\\ 2) high-def \textbf{VR} on iPad\\ 3) \textbf{VR} on iPhone\end{tabular} & \begin{tabular}[c]{@{}l@{}}1) 
                Webcam in MacBook Air\\ 2) \textbf{FC} of Google Nexus5 Mob 
            \end{tabular}                     & \begin{tabular}[c]{@{}l@{}}train\\ test\end{tabular}         & \begin{tabular}[c]{@{}l@{}}90\\ 120\end{tabular}        & \begin{tabular}[c]{@{}l@{}}30\\ 84\end{tabular}         & \begin{tabular}[c]{@{}l@{}}120\\ 204\end{tabular}       \\
            
            \hline
            Replay-Mobile & \begin{tabular}[c]{@{}l@{}}1) \textbf{PR} (A4)\\ 2) \textbf{VR} on matte-screen\end{tabular}                                & \begin{tabular}[c]{@{}l@{}}1) \textbf{FC} of iPad Mini2 Tablet
                \\ 2) \textbf{FC} of  LG-G4 Mobile
            \end{tabular} & \begin{tabular}[c]{@{}l@{}}train\\ devel\\ test\end{tabular} & \begin{tabular}[c]{@{}l@{}}192\\ 256\\ 192\end{tabular} & \begin{tabular}[c]{@{}l@{}}120\\ 160\\ 110\end{tabular} & \begin{tabular}[c]{@{}l@{}}312\\ 416\\ 302\end{tabular}
            \\
            \hline
            \multicolumn{1}{l}{} & \multicolumn{6}{l}{\textbf{FC}: Front-Camera, 
                \textbf{PR}: Hard-copy print of high-res photo, \textbf{VR}: Video replay}
        \end{tabular}
    \end{center}
\end{table*}

\begin{table*}[!ht]
    \centering
    \caption{Results of Proposed DC-guided-DA for Face-PAD in ACER\% at threshold $0.5$.}
    \label{tab:face-da-all}
    \small
    \begin{tabular}{|l|cc|cc|cc|r|}
        \hline
        train$\rightarrow$test             & RA$\rightarrow$M & RA$\rightarrow$RM & M$\rightarrow$RA  & M$\rightarrow$RM  & RM$\rightarrow$RA & RM$\rightarrow$M & Average \\
        \hline
        Source-only                  & 34   & 49.8  & 39.4  & 15.6  & 42.3  & 42   &   37.18
        \\
        
        DA w/o clustering & 29.6 & 47.2  & 49.25 & 11.35 & 45    & \textbf{2.9} &  30.88
        \\
        
        DCDA w/o $L^t_{\tilde{y}}$ & 18.35 & 49.2 & 10.40 & 2.25 & 11.65 & 37.80 & 19.94
        \\
        DCDA & \textbf{0} & \textbf{0}  & \textbf{0.15}  & \textbf{1.6}  & \textbf{1.15}   & \textbf{1.65} & \textbf{0.76}
        \\
        \hline
        
        \multicolumn{1}{l}{} & \multicolumn{7}{l}{\small\textbf{RA}: Replay-Attack, \textbf{M}: MSU-MFSD, \textbf{RM}: Replay-Mobile}
    \end{tabular}
\end{table*}

\begin{table*}
	\centering
	\caption{Comparison with SOTA in HTER\%.}
	\label{tab:face-da-comp}
	\begin{tabular}{|l|cc|r|}
		\hline
		& RA$\rightarrow$M & M$\rightarrow$RA  & Average \\
		\hline
		KSA$^\S$~\cite{UDA_face_journal_2018} & 18.6$^\star$ & 23.3$^\star$ & 20.95 \\
		ADA~\cite{impr_face_ADA_2019} & 30.5 & 5.1 & 17.8 \\
		PAD-GAN~\cite{cross_dom_face_DA_2020} & 23.2 & 8.7 & 15.95 \\
		SSDG~\cite{Jia_2020_CVPR_SSDG} & 7.38$^{\star\star}$ & 11.7$^{\star\star}$ & 9.54 \\
		DCDA (Proposed) & \textbf{0} & \textbf{0.15} & \textbf{0.08}
		\\
		\hline
		\multicolumn{4}{l}{\small{$^\star$ On concatenated CoALBP and LPQ features in HSV and YCbCr color space}}\\
		
		\multicolumn{4}{l}{\small{$^{\star\star}$ Source-domain includes two other datasets}}
		
	\end{tabular}
\end{table*}

\begin{figure*}[!h]
	\centering
	
	\begin{subfigure}[b]{0.48\textwidth}
		\begin{subfigure}[b]{0.49\textwidth}
			\centering
			\includegraphics[width=\textwidth]{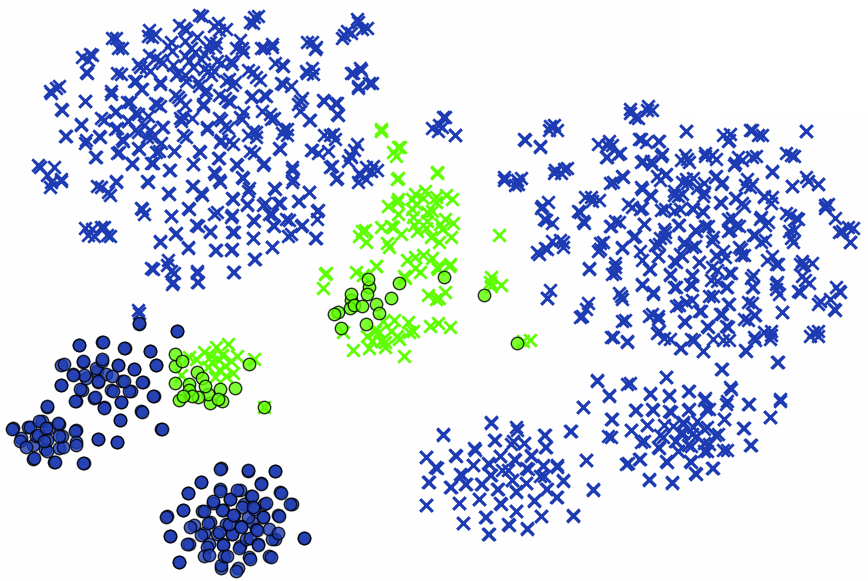}
			\caption{RA$\rightarrow$M$^{(-)}$}
		\end{subfigure}
		\hfill
		\begin{subfigure}[b]{0.49\textwidth}
			\centering
			\includegraphics[width=\textwidth]{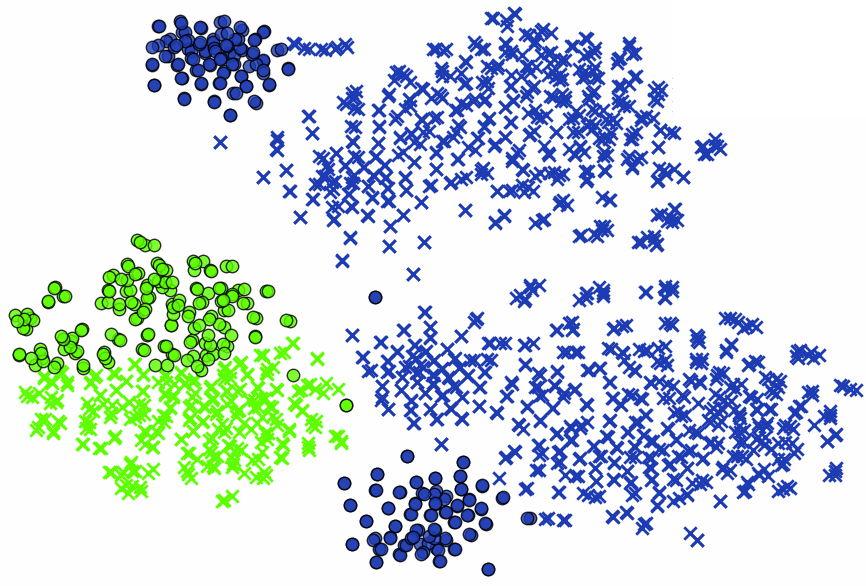}
			\caption{RA$\rightarrow$RM$^{(-)}$}
		\end{subfigure}
		\\
		
		\begin{subfigure}[b]{0.49\textwidth}
			\centering
			\includegraphics[width=\textwidth]{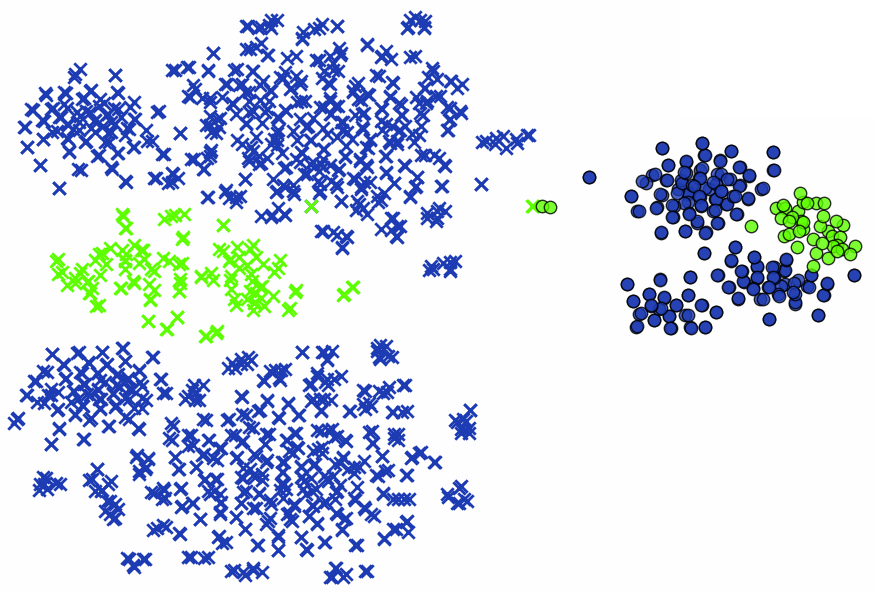}
			\caption{RA$\rightarrow$M$^{(\star\star)}$}
			\label{fig:ra_m}
		\end{subfigure}
		\hfill
		\begin{subfigure}[b]{0.49\textwidth}
			\centering
			\includegraphics[width=\textwidth]{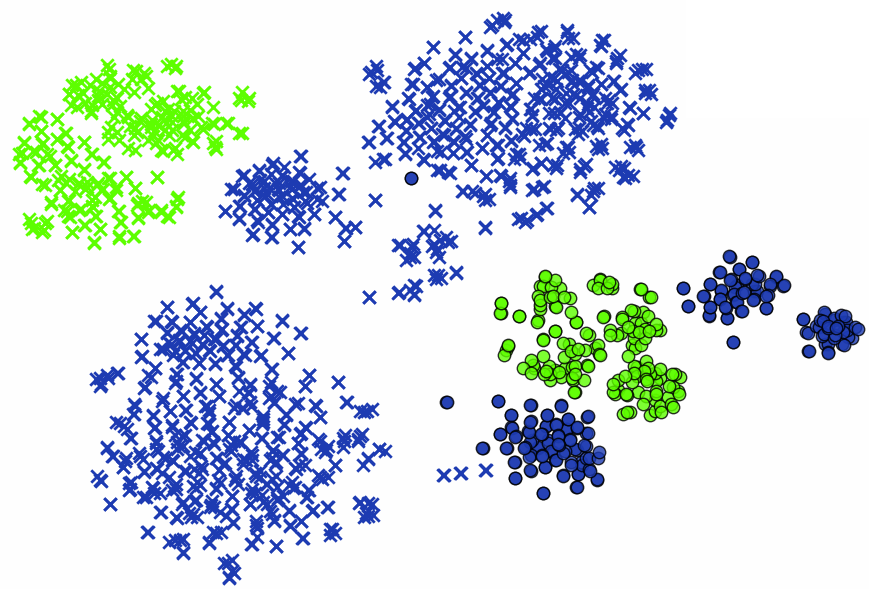}
			\caption{RA$\rightarrow$RM$^{(\star\star)}$}
			\label{fig:ra_rm}
		\end{subfigure}   
	\end{subfigure}
	\hfill
	\begin{subfigure}[b]{0.48\textwidth}
		\begin{subfigure}[b]{0.49\textwidth}
			\centering
			\includegraphics[width=\textwidth]{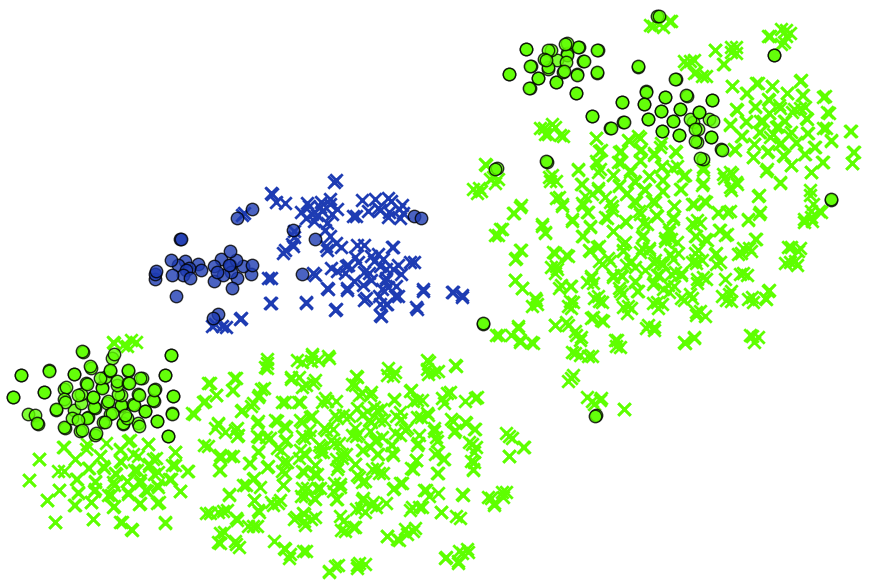}
			\caption{M$\rightarrow$RA$^{(-)}$}
		\end{subfigure}
		\hfill
		\begin{subfigure}[b]{0.49\textwidth}
			\centering
			\includegraphics[width=\textwidth]{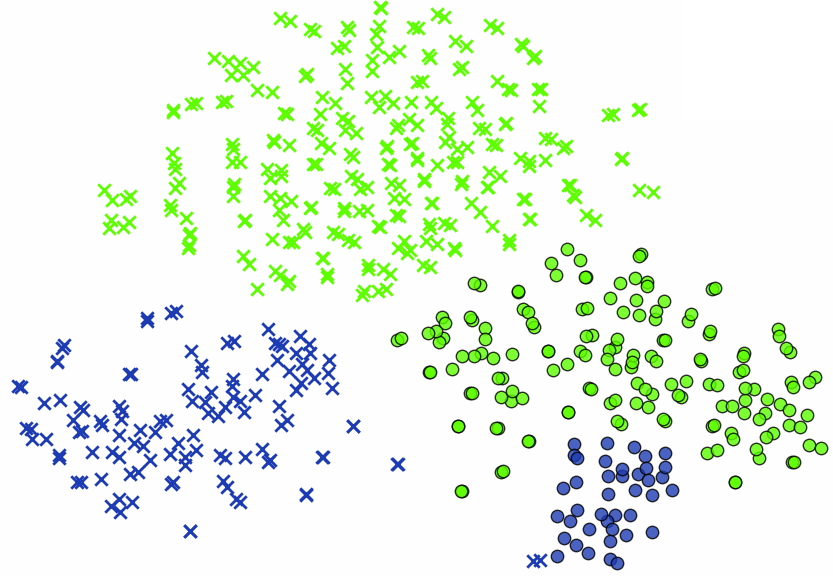}
			\caption{M$\rightarrow$RM$^{(-)}$}
		\end{subfigure}
		\\
		\begin{subfigure}[b]{0.49\textwidth}
			\centering
			\includegraphics[width=\textwidth]{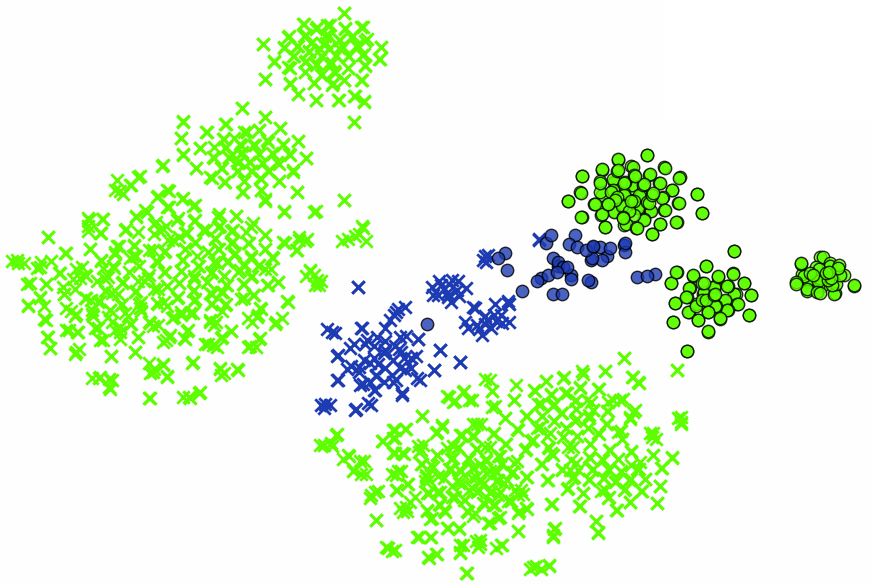}
			\caption{M$\rightarrow$RA$^{(\star\star)}$}
			\label{fig:m_ra}
		\end{subfigure}
		\hfill
		\begin{subfigure}[b]{0.49\textwidth}
			\centering
			\includegraphics[width=\textwidth]{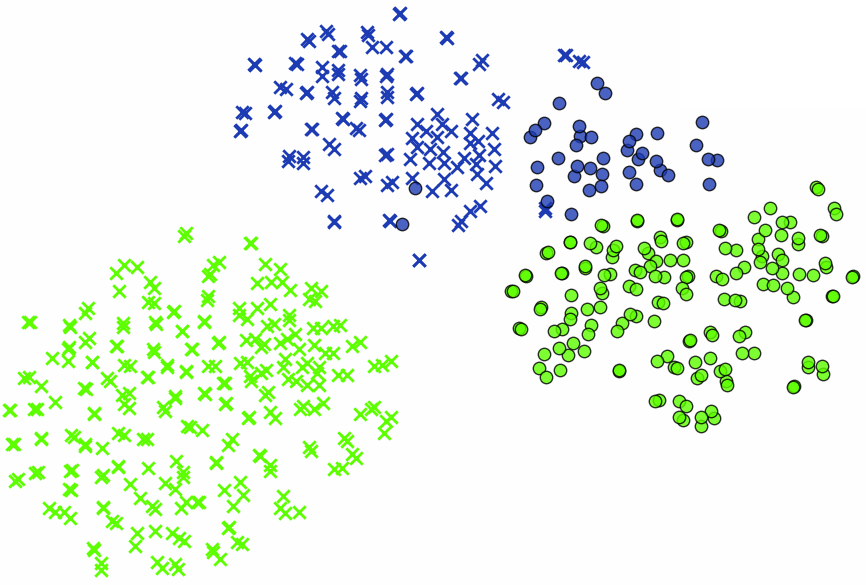}
			\caption{M$\rightarrow$RM$^{(\star\star)}$}
		\end{subfigure}
		
		\label{fig:tsne2}
	\end{subfigure}
	
	\begin{subfigure}[b]{0.5\textwidth}
		\centering
		
		\begin{subfigure}[b]{0.49\textwidth}
			\centering
			\includegraphics[width=\textwidth]{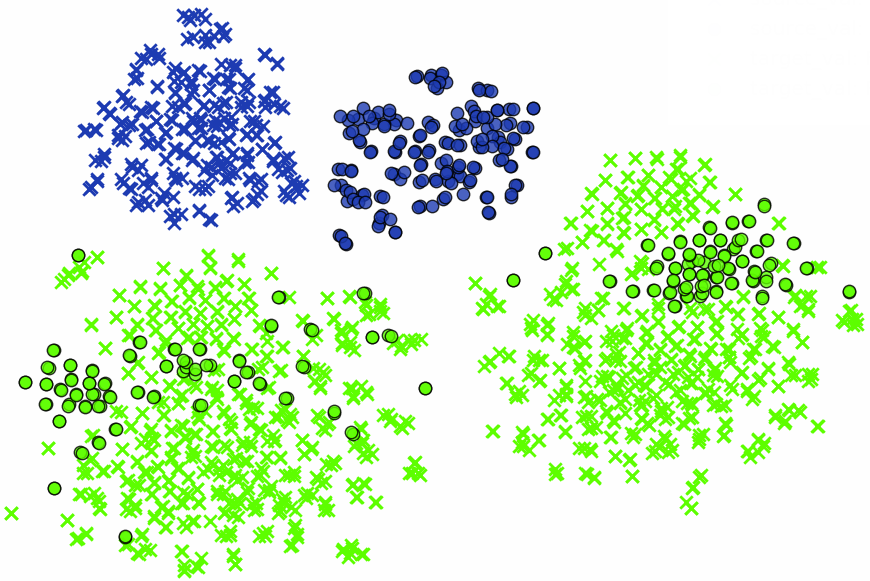}
			\caption{RM$\rightarrow$RA$^{(-)}$}
		\end{subfigure}
		\hfill
		\begin{subfigure}[b]{0.49\textwidth}
			\centering
			\includegraphics[width=\textwidth]{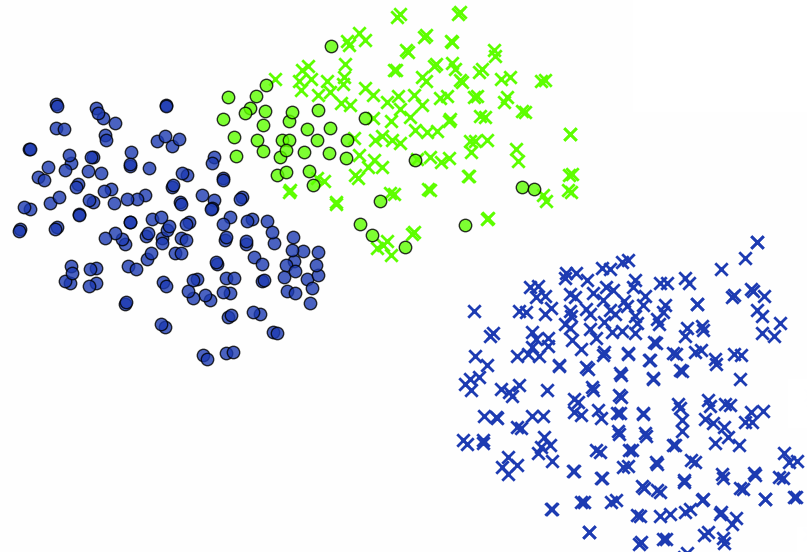}
			\caption{RM$\rightarrow$M$^{(-)}$}
		\end{subfigure}
		\\
		
		
		\begin{subfigure}[b]{0.49\textwidth}
			\centering
			\includegraphics[width=\textwidth]{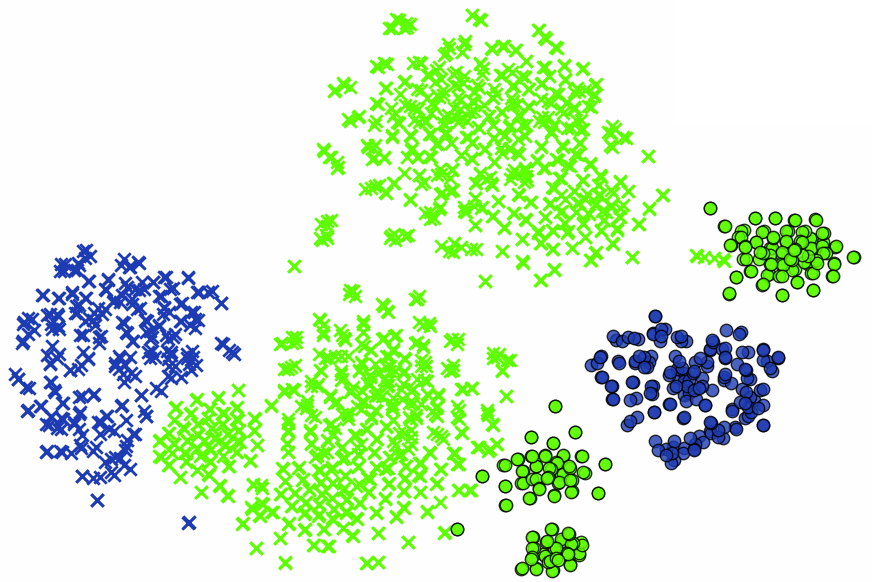}
			\caption{RM$\rightarrow$RA$^{(\star\star)}$}
			\label{fig:rm_ra}
		\end{subfigure}
		\hfill
		\begin{subfigure}[b]{0.49\textwidth}
			\centering
			\includegraphics[width=\textwidth]{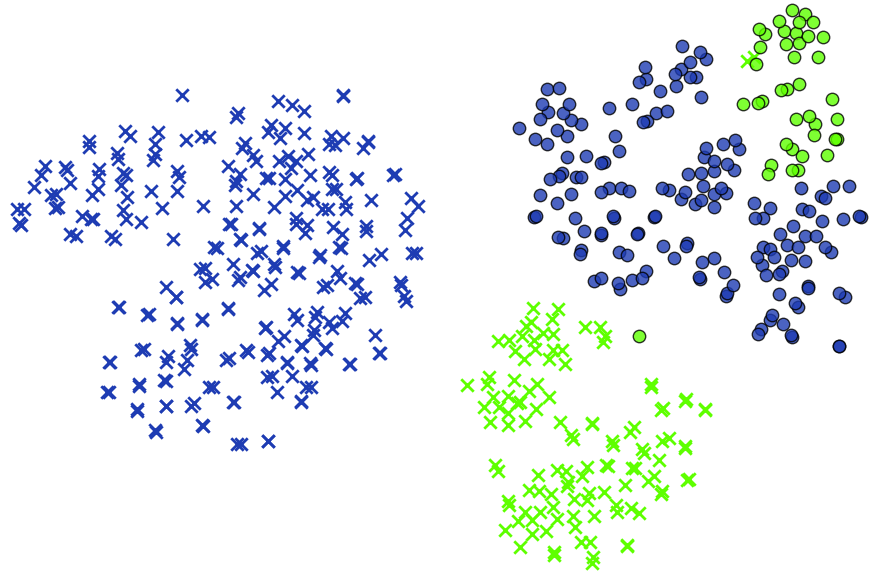}
			\caption{RM$\rightarrow$M$^{(\star\star)}$}
		\end{subfigure}
	\end{subfigure}
	
	\caption{t-SNE visualization analysis. Upper row $^{(-)}$: DA without clustering, Bottom row $^{(\star\star)}$: Proposed DC-guided-DA. 
		$Blue$: Source, $Green$: Target, $\circ$: Bona-fide, $\times$: Attack. Best viewed in color.}
	\label{fig:tsne}
\end{figure*}

\section{\uppercase{Experiments and results}}\label{sec:experiments}

\subsection{Face PAD datasets}

Table~\ref{tab:face-db-numbers} summarizes the total number of samples present in each subset of the datasets used, in addition to the Presentation Attack Instruments (PAI) used and the sensors used in recording videos for authentication.

\textbf{Replay-Attack}~\cite{547_12_f} is one of the earliest datasets presented in literature for the problem of face spoofing 
It consists of 1200 short videos from 50 different subjects with resolution $320 \times 240$ from 50 different subjects. 
Attack scenario include "hard-copy print-attack", "mobile-photo attack" and "high-definition screen attack". Attacks are presented to the sensor (regular webcam) either with a "fixed" tripod, or by an attacker holding the presenting device (printed paper or replay device) with his/her "hand".

\textbf{MSU Mobile Face Spoofing Database (MSU-MFSD)}~\cite{515_15_f}  
targets the problem of face spoofing on smartphones
. The dataset includes real and spoofed videos from 35 subjects
. Two devices were used, the 
webcam of a MacBook Air 
with resolution $640 \times 480$ and the front facing camera of 
a smartphone with $720 \times 480$ resolution. 
Three attack scenarios are used: print-attack on A3 paper, video replay attack on the screen of an iPad and video replay attack on a smartphone.

\textbf{Replay-Mobile}~\cite{564_16_f} was released by the same research institute that released Replay-Attack. It has 1200 short videos from 40 subjects captured by two mobile devices at resolution $720 \times 1280$. Each subject has ten bona-fide accesses and 16 attack videos under different attack modes. 
Two types of attack are present: photo-print and matte-screen attack displaying digital-photo or video. 

\subsection{Experimental setup}

Our experiments were performed on NVIDIA GeForce 840m GPU with CUDA version 11.0. Bob package~\cite{bob2012} was used for datasets management and PyTorch was used for models and training.
Evaluation metrics for PAD are the ISO/IEC 30107-3:2017\footnote{\url{https://www.iso.org/standard/67381.html}} metrics. Attack Presentation Classification Error Rate (APCER), Bona-fide Presentation Classification Error Rate (BPCER) and their Average Classification Error Rate (ACER) ($(APCER + BPCER) / 2$) is used for reporting results in the tables.

\subsection{Results and Discussion}

Table~\ref{tab:face-da-all}
presents results of our proposed DC-guided UDA for face PAD on 
the 3 benchmark face datasets used.
Results are reported as the average ACER \% of three runs, ACER is calculated on the test subset of the target dataset. The first row 
represents the results obtained by fine-tuning a MobileNetV3 classification network on source dataset only without domain adaptation. 
We performed experiments to study the influence of each model component 
on the overall performance of the algorithm. Clustering components and losses were removed and only Domain Adaptation was performed, results in the second row of Table~\ref{tab:face-da-all} show only slight improvement over source-only trained models. Then, adding clustering components with target psuedo-labels estimation and target clustering loss $L^t_{cl}$, but without updating the classifier $\mathcal{C}$ with target classification loss $L^t_{\tilde{y}}$, yielded a significant decrease in the target classification error on most datasets as shown in third row. However, though feature extraction network is trying to learn domain-invariant features, the classifier trained on source-samples only still fails in some cases to achieve low errors on some target datasets. For example, the classifier trained on Replay-Attack dataset fails to discriminate the attack and bona-fide samples on Replay-Mobile dataset. 

Finally, the last row shows results obtained by our full proposed DCDA framework, which achieves near-perfect classification of the unlabeled target samples. 
Comparison with state-of-the art DA-based face PAD solutions is provided in Table~\ref{tab:face-da-comp} showing superiority of our proposed DC-guided-DA framework.
Furthermore, $t$-SNE visualization analysis is presented in Figure~\ref{fig:tsne}, comparing our proposed architecture, with models trained using Domain Adaptation only. The visualizations show that our proposed framework could align the classification boundaries for both source and target datasets, it also shows the diversity of attack and sensors types present in the same dataset that form clusters in the same class of the same dataset, for example Replay-Attack in Figure~\ref{fig:tsne} parts~\ref{fig:ra_m}, ~\ref{fig:ra_rm}, ~\ref{fig:m_ra} and~\ref{fig:rm_ra}.

\section{\uppercase{Conclusion and Future work}}\label{sec:conc}

In this paper, we proposed an approach that exploits unsupervised adversarial domain adaptation guided with target clustering, in order to improve the generalization ability for face PAD. Specifically, our framework utilizes UDA to learn domain invariant features that could leverage from the labeled source samples to classify the unlabeled samples from target domain. Yet, the approach succeeds to preserve the intrinsic properties of the target domain via deep clustering of target embedding features. 
Our approach is trained in an end-to-end fashion and succeeds to reach perfect adaptation to the target domain when evaluated on public benchmark datasets, reaching only 0 - 2\% cross-dataset error. 
Our future work would focus on evaluating on more variable datasets, 
in addition to reducing the dependency of the model during training on target domain samples from both classes, trying to let the model focuses on learning from bona-fide samples with minimal attack samples contribution.

\bibliographystyle{apalike}
{\small
    \bibliography{DCDA}
}

\end{document}